\documentclass[]{article}
\usepackage{proceed2e}
\PassOptionsToPackage{numbers}{natbib}
\usepackage{graphicx}
\usepackage[font=small,labelfont=bf]{caption}
\usepackage{subcaption}
\usepackage{amssymb}
\usepackage{amsmath}
\usepackage{amsfonts}
\usepackage[bottom]{footmisc}

\usepackage{algorithm}
\usepackage{algorithmic}

\usepackage{amsthm}
\usepackage{siunitx}
\usepackage{array}

\usepackage{placeins}
\usepackage{hyperref}
\usepackage{url}
\usepackage{xcolor}
\usepackage{multirow}

\usepackage{natbib}

\usepackage{times}
\newcommand{\dataset}[0]{ \mathcal{D} }

\title{Probabilistic Safety for Bayesian Neural Networks}

\author{ {\bf Matthew Wicker\thanks{\hspace*{0.15cm}Equal Contributions.}} \\
University of Oxford \\
\And
{\bf Luca Laurenti}$^*$  \\
University of Oxford \\
\And
{\bf Andrea Patane}$^*$  \\
University of Oxford \\
\And
{\bf Marta Kwiatkowska}  \\
University of Oxford \\
}

\begin{document}

\maketitle

\newtheorem{theorem}{Theorem}
\newtheorem{definition}{Definition}
\newtheorem{problem}{Problem}
\newtheorem{myexam}{Example}
\newtheorem{assumption}{Assumption}
\newtheorem{proposition}{Proposition}
\newtheorem{remark}{Remark}
\newtheorem{lemma}{Lemma}
\newtheorem{corollary}{Corollary}

\newcommand{\MRW}[1]{{\color{blue}[MW: #1]}}
\newcommand{\ap}[1]{{\color{orange}[AP: #1]}}
\newcommand{\LL}[1]{{\color{red}[LL: #1]}}
\newcommand{\MK}[1]{{\color{green}[MK: #1]}}

\newcommand{\pX}{\mathbf{x}}

\newcommand{\BNN}[0]{{f}^{\mathbf{w}}(x)}

\begin{abstract}
We study probabilistic safety for Bayesian Neural Networks (BNNs) 
 under adversarial input perturbations.
Given a compact set of input points, $T \subseteq \mathbb{R}^m$, we study the probability w.r.t.\ the BNN posterior that all the points in $T$ are mapped to the same 
region $S$ in the output space.
In particular, this can be used to evaluate the probability that a network sampled from the BNN is vulnerable to adversarial attacks.
We rely on relaxation techniques from non-convex optimization to develop a method 
for computing a lower bound  
on probabilistic safety for BNNs,
deriving explicit procedures for the case of interval and linear function propagation techniques.
We apply our methods to BNNs trained on a regression task, airborne collision avoidance, and MNIST, empirically showing that our approach allows one to certify probabilistic safety of BNNs with millions of parameters.

\end{abstract}

\section{INTRODUCTION}
%
Although Neural Networks (NNs) have recently achieved state-of-the-art performance 
\cite{he2015delving}, 
they are 
susceptible to several vulnerabilities \cite{biggio2018wild}, with adversarial examples being one of the most prominent among them \cite{IntruigingProperties}.
Since adversarial examples are arguably intuitively related to \emph{uncertainty} \cite{kendall2017uncertainties}, Bayesian Neural Networks (BNNs), i.e. NNs with a probability distribution placed over their weights and biases \cite{neal2012bayesian}, have recently been proposed as a potentially more robust learning paradigm \cite{carbone2020robustness}. 
While retaining the advantages intrinsic to deep learning (e.g.\ representation learning), BNNs also enable principled evaluation of model uncertainty, which can be taken into account at prediction time to enable safe decision making.

Many techniques have been proposed  for the evaluation of the robustness of BNNs, including generalisation of NN gradient-based adversarial attacks to BNN posterior distribution \cite{liu2018adv}, statistical verification techniques for adversarial settings \cite{cardelli2019statistical}, as well as pointwise (i.e.\ for a specific test point $x^*$) uncertainty evaluation techniques \cite{smith2018understanding}. 
However, to the best of our knowledge, methods that formally (i.e., with certified bounds) give guarantees on the behaviour of BNNs against adversarial input perturbations in probabilistic settings are missing.

In this paper we aim at 
analysing \emph{probabilistic safety} for BNNs.
Given a BNN, a set $T \subseteq \mathbb{R}^m$ in the input space, and a set $S\subseteq \mathbb{R}^{n_c}$ in the output space, probabilistic safety is defined as the probability that for all $x\in T$ the prediction of the BNN is in $S$. 
In adversarial settings, the input region $T$ is built around the neighborhood of a given test point $x^*$, while the output region $S$ is defined around the BNN prediction on $x^*$, so that probabilistic safety translates into computing the probability that adversarial perturbations of $x^*$ cause small variations in the BNN output. 
Note that probabilistic safety represents a probabilistic variant of the notion of safety commonly used to certify deterministic NNs 
\cite{ruan2018reachability}.

Unfortunately, computation of probabilistic safety for a BNN over a compact set $T$ and for an output region $S$ is not trivial, as it involves computing  the probability that a deterministic NN sampled from the BNN posterior is safe (i.e., all the points in a given set are mapped by the network to a given output set), which is known to be NP-complete 
\cite{katz2017reluplex}. 
Nevertheless, we derive a method for the computation of a \emph{certified lower bound} for probabilistic safety.
In particular,  we show that the computation of probabilistic safety for BNNs is equivalent to computing the measure, w.r.t.\ BNN posterior, of the set of weights for which the resulting deterministic NN is safe, i.e., robust to adversarial perturbations.
We compute a subset of such weights, $\hat{H}$, and build on relaxation techniques from non-linear optimisation to check whether, given a compact set $T$ and a safe output region $S$, all the networks instantiated by weights in $\hat{H}$ are safe.
We provide lower bounds for the case of Interval Bound Propagation (IBP) and Linear Bound Propagation (LBP) for BNNs trained with variational inference (VI) \cite{blundell2015weight}. However, we note that the derived bounds can be extended to other approximate Bayesian inference techniques. 

We experimentally investigate the behaviour of our method on a regression task, the VCAS dataset and the MNIST dataset, for a range of properties and different BNN architectures, and applying bounds obtained both via IBP and LBP.
We show that our method allows one to compute non-trivial, certified lower bounds on the probabilistic safety of BNNs with millions of weights in a few minutes. 

In summary, this paper makes the following main contributions.\footnote{An implementation to reproduce all the experiments can be found at: \url{https://github.com/matthewwicker/ProbabilisticSafetyforBNNs}.  }
\begin{itemize}
    \item  We propose 
    a framework based on relaxation techniques from non-convex optimisation
    for the 
    analysis of 
    probabilistic safety for BNNs with general activation functions and  multiple hidden layers.
    \item We derive explicit procedures based on IBP and LBP for the computation of the set of weights for which the corresponding NN sampled from the BNN posterior distribution is safe.
    \item  On various datasets we show that our algorithm can efficiently verify multi-layer BNNs with millions of parameters in a few minutes.
\end{itemize}

\paragraph{Related Work.}
Most existing certification methods 
are designed for deterministic neural networks (see e.g., \cite{katz2017reluplex}) and cannot be used for verification of Bayesian models against probabilistic properties. In particular, in \cite{weng2018towards,zhang2018efficient,gowal2018effectiveness}  Interval Bound Propagation (IBP) and Linear Bound Propagation (LBP) approaches have been employed for certification of deterministic neural networks. 
However, these methods cannot be used for BNNs because they all assume that the weights of the networks are deterministic, i.e\ fixed to a given value, while in the Bayesian setting we need to certify the BNN for a continuous range of values for weights that are not fixed, but distributed according to the BNN posterior.
We extend IBP and LBP to BNNs in Section \ref{sec:IBPandLBP}. 

Bayesian uncertainty estimates have been shown to  
empirically flag adversarial examples in \cite{rawat2017adversarial, smith2018understanding}. 
However, these approaches only consider pointwise uncertainty estimates, that is, specific to a particular test point.
In contrast, probabilistic safety aims at  computing the uncertainty for compact subspaces of 
input points, thus taking into account worst-case adversarial perturbations of the input point when considering its neighbourhood.
A probabilistic property analogous to that considered in this paper has been studied for BNNs in \cite{cardelli2019statistical, michelmore2019uncertainty}.
However,  the solution methods presented in \cite{cardelli2019statistical, michelmore2019uncertainty} are statistical, with confidence bounds,  and in practice cannot give certified guarantees for the computed values, which are important for safety-critical applications.
Our approach, instead building on non-linear optimisation relaxation techniques, computes a certified lower bound.

Methods to compute probabilistic adversarial measures of robustness in Bayesian learning settings have been explored for Gaussian Processes (GPs),  both for regression \cite{cardelli2018robustness} and classification tasks \cite{smith2018understanding,blaas2019robustness}.
However, the vast majority of inference methods employed for BNNs do not have a Gaussian approximate distribution in latent/function space, even if they assume a Guassian distribution in the weight space \cite{blundell2015weight}.
Hence, certification techniques for GPs cannot be directly employed for BNNs. 
In fact, because of the non-linearity of the BNN structure, even in this case the distribution of the BNN is not Gaussian in function space. 
Though methods to approximate BNN inference with GP-based inference have been proposed \cite{emtiyaz2019approximate}, the guarantees obtained in this way would apply to the approximation and not the actual BNN, and
would not provide error bounds.
In contrast, our method is specifically tailored to take into account the non-linear nature of BNNs and can be directly applied to a range of approximate Bayesian inference techniques.

\section{BAYESIAN NEURAL NETWORKS (BNNs)} 
In this section we review BNNs. 
We write $\BNN = [f_1^{\mathbf{w}}(x),\ldots,f_{n_c}^{\mathbf{w}}(x)]$ to denote a BNN with $n_c$ output units and an unspecified number of hidden layers, where ${\mathbf{w}}$ is the weight vector random variable.
Given a distribution over ${\mathbf{w}}$ and $w \in \mathbb{R}^{n_w}$, a weight vector sampled from the distribution of ${\mathbf{w}}$, we denote with ${f}^w(x)$ the corresponding deterministic neural network with weights fixed to $w$.
Let $ \dataset =\{(x_i,y_i), i \in \{1,...,N_{\mathcal{D}} \} \}$ be the training set. 
In Bayesian settings, we assume a prior distribution over the weights, i.e.\ $ {\mathbf{w}} \sim p(w)$, 
so that learning amounts to computing the posterior distribution over the weights, $p (w \vert \dataset )$, via the application of Bayes rule. 
Unfortunately, because of the non-linearity introduced by the neural network architecture, the computation of the posterior cannot be done analytically \cite{mackay1992practical}.

In this work we focus on Gaussian Variational Inference (VI) approximations via Bayes by Backprop  \cite{blundell2015weight}.
In particular, VI proceed by finding a suitable Gaussian approximating distribution $q(w) = \mathcal{N}(w|\mu,\Sigma) $ for the posterior distribution, i.e.\ such that $ q(w) \approx p(w \vert \dataset)$. 
The core idea is that the mean and covariance of  $q(w)$ are
iteratively optimized by minimizing a divergence measure between $q (w)$ and $ p (w | \dataset)$. 
In particular, in Section \ref{sec:methodology} we give explicit bounds for BNN certification for variational inference; however, we remark that the techniques developed in this paper  also extend to other classes of distributions and approximate inference methods, such as Hamiltonian Monte Carlo \cite{neal2012bayesian} or dropout  \cite{gal2016dropout}, with the caveat that in those cases the integral 
in Eqn.\ \eqref{Eq:IntegralSafety}  (Section \ref{sec:methodology}) will not be Gaussian  and will need to be computed by means of Monte Carlo sampling techniques.

\section{PROBLEM FORMULATION}
\begin{figure}[h]
    \centering
    \includegraphics[width=0.45\textwidth]{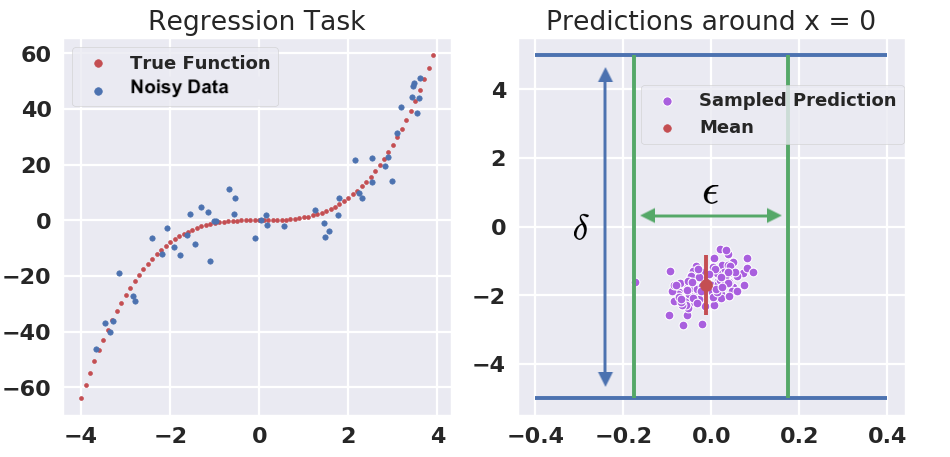}
    \caption{Left: 50 points sampled uniformly from $[-4,4]$ and their corresponding outputs according to $y = x^3 + \mathcal{N}(0,5)$. Right:
    For $T=[-\epsilon, \epsilon]$ and $S=[-\delta, \delta]$, with $\epsilon =0.2$ and $\delta = 5$, we visualize the property $P_{\text{safe}}(T,S)$. The red dot and bar  depicts the mean and the standard deviation of the BNN prediction in $x^* = 0$.}
    \label{fig:toyexample1}
\end{figure}

A BNN is a stochastic process whose randomness comes from the weight distribution. 
Therefore, in order to study 
robustness, its probabilistic nature should be considered. 
In this paper we focus on probabilistic safety, which is a widely employed measure to characterize the robustness of stochastic models \cite{abate2008probabilistic,9003411}, and also represents a probabilistic generalization of the notion of safety commonly used to guarantee the robustness of deterministic neural networks  against adversarial examples \cite{ruan2018reachability}.
In particular,  probabilistic safety for BNNs is defined as follows.
\begin{definition}\label{Def:ProbSafety}
Let $f^{\mathbf{w}}$ be a BNN, $\mathcal{D}$ a dataset, $T\subset \mathbb{R}^m$ a compact set of input points, and $S\subseteq \mathbb{R}^{n_c}$ a safe set. Then, probabilistic safety is defined as 
\begin{align}
\label{Eq:ProbSafety}
    P_{\text{safe}}(T,S):=Prob_{w \sim \mathbf{w}}(\forall x \in T, f^{{w}}(x) \in S | \mathcal{D}).
\end{align}
\end{definition}
$P_{\text{safe}}(T,S)$ is the probability that for all input points in $T$ the output of the BNN belongs to a given output set $S$. 
Note that the probabilistic behaviour in $P_{\text{safe}}(T,S)$ is uniquely determined from the distribution over the weights random variable $\mathbf{w}$. In particular, no distribution is assumed in the input space. Hence, $P_{\text{safe}}(T,S)$ represents a probabilistic measure of robustness.
We make the following assumption on $S$.
\begin{assumption}\label{assumption}
We assume that ${S}$ is described by $n_S$ linear constraints on the values of the final layer of the BNN, that is, 
$$ S=\{y\in \mathbb{R}^{n_c} \, |  \, C_S y + d_S \geq 0, C_S \in \mathbb{R}^{n_S\times n_c},d_S\in \mathbb{R}^{n_S}   \}  $$
\end{assumption}
We stress that, as discussed in \cite{gowal2018effectiveness}, this assumption 
encompasses most properties of interest for the verification of neural networks. 
We refer to $C_S y + d_S \geq 0$ as the \textit{safety specification} associated to the safe set $S$.
Below, in Example \ref{ex:Running}, we illustrate the notion of probabilistic safety on an example.
\begin{myexam}
\label{ex:Running}
We consider a regression task where we learn a BNN from noisy data centred around the function $y=x^3$, as illustrated in Figure \ref{fig:toyexample1}. We  let $T = [-\epsilon, \epsilon]$ and $S = [-\delta,\delta]$, with $\epsilon = 0.2$ and $\delta = 5$ be two intervals. Then, we aim at computing $P_{\text{safe}}(T,S)$, that is, the probability that for all $x\in [-\epsilon, \epsilon]$, $f^{\mathbf{w}}(x)\in [-\delta,\delta]$. 
\end{myexam}

Probabilistic safety can be used in a regression setting to formally account for the uncertainty in the learned model. For instance, it can be
employed 
in a model-based reinforcement learning scenario to ensure the safety of the learned model under uncertain environments \cite{berkenkamp2017safe}.
In a classification problem, $P_{\text{safe}}(T,S)$ could be used to evaluate the uncertainty around the model predictions  in adversarial settings \cite{gal2016uncertainty}.
We remark that probabilistic safety is also related to other adversarial robustness measures in the literature  \cite{dvijotham2018verification,blaas2019robustness}.
In particular, it is straightforward to show (see Proposition 4 
in the Supplementary Material in \cite{wicker2020probabilistic}) that
$$ P_{\text{safe}}(T,S) \leq \inf_{x \in T} Prob_{w\sim \mathbf{w}}(f^{{w}}(x)\in S). $$
Moreover, if for $i \in \{1,...,n_c \}$ and $a\in \mathbb{R}_{>0}$ we assume that $S=\{y\in \mathbb{R}^{n_c}\, |\, y_{i}>a  \}$, then it holds that
$$ a P_{\text{safe}}(T,S) \leq  {\inf_{x \in T} \mathbb{E}_{w \sim {\mathbf{w}}}[ f^{w}_i(x)  ]} .  $$


\paragraph{Approach Outline}
Probabilistic safety, $P_{\text{safe}}(T,S)$, is not trivial to compute for a given compact set $T$ in the input space and safe set $S$ that satisfies Assumption \ref{assumption}.
In fact, already in the case of deterministic neural networks, safety has been shown to pose an NP-complete problem \cite{zhang2018efficient}.
Therefore, in Section \ref{sec:methodology} we derive a method for lower-bounding (i.e., for the worst-case analysis) of  $P_{\text{safe}}(T,S)$, which can be used for certification of the probabilistic safety of BNNs. 
We first show that the computation of $P_{\text{safe}}(T,S)$ is equivalent to computing the maximal set of weights $H$  such that the corresponding deterministic neural network is safe, i.e., $H=\{w\in \mathbb{R}^{n_w} | \forall x \in T, f^w(x)\in S \}$.
The computation of $H$ is unfortunately itself not straightforward. 
Instead, in Section \ref{sec::Algorithm}, we design a method to compute a subset of safe weights $\hat{H} \subseteq H$, and discuss how $\hat{H}$ can be used to compute a certified lower bound to $P_{\text{safe}}(T,S)$.
%
Our method (detailed in Algorithm \ref{alg:simplereachability}) works by sampling weights $w$ from the posterior BNN distribution and iteratively building safe weight regions, in the form of disjoint hyper-rectangles, around the sampled weights. 
This requires a subroutine that, given $\hat{H}$, checks whether it constitutes a safe set of weights or not, that is, verifies if the statement $\hat{H} \subseteq H$ is true.
In Section \ref{sec:IBPandLBP}, we derive two alternative approaches for the solution of this problem, one based on Interval Bound Propagation (IBP) (introduced in Section \ref{sec:IBP}) and the other on Linear Bound Propagation (LBP)  (introduced in Section \ref{sec:LBP}).
These work by propagating the input region $T$ and the weight rectangle $\hat{H}$ through the neural network, in the form of intervals (in the case of IBP) and lines (in the case of LBP) and checking whether the resulting output region is a subset of $S$, thus deciding if $\hat{H}$ is a safe set of weights or not.
%
%

\begin{figure*}[h]
    \centering
    \includegraphics[width=0.80\textwidth]{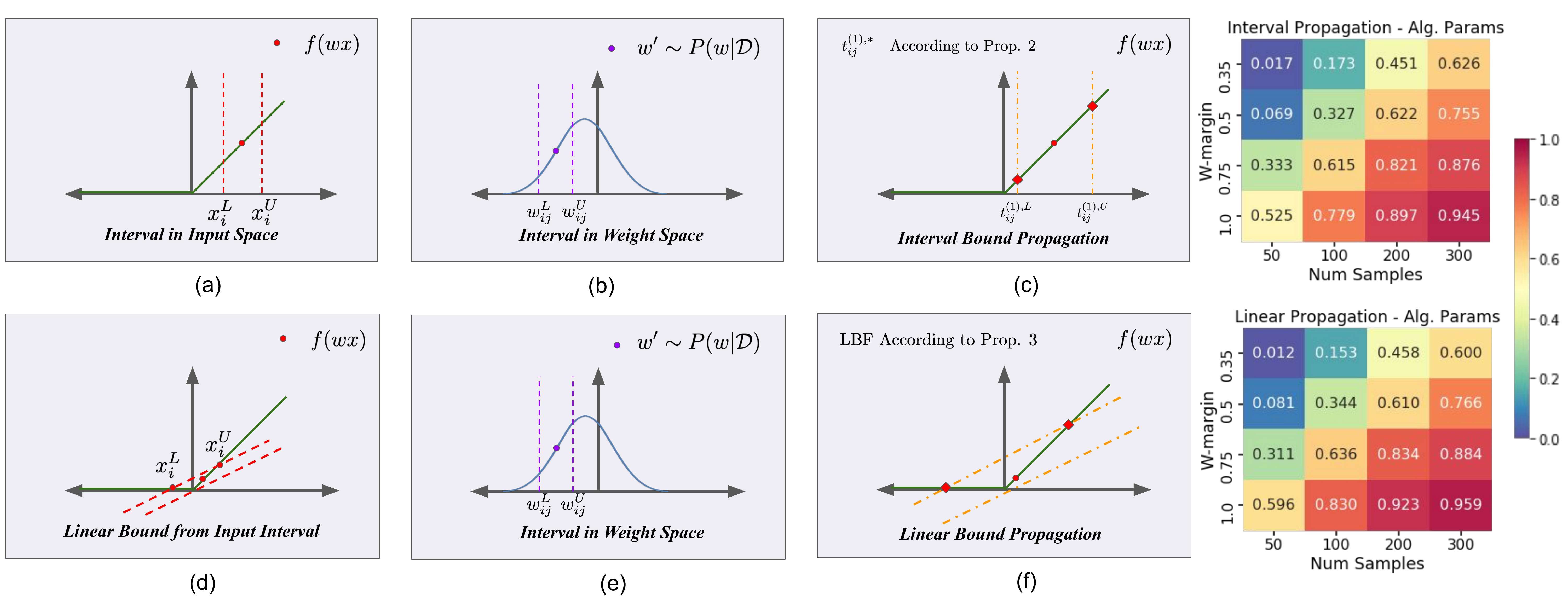}
    \caption{Example of IBP (first row) and LBP (second row) for BNNs. For IBP we consider an interval in the input space (a) together with an interval in the weight space (b). The two intervals are combined to obtain an interval in the output, which is guaranteed to contain the network output (c).
    In the LBP case, the input bound and the weight interval are combined to obtain linear bounds that contain the network output at any layer (d)-(e)-(f). In the last column we show the application of Algorithm \ref{alg:simplereachability} for the computation of the property illustrated in Example \ref{ex:Running}. We consider different values of the parameters required by the algorithm for both IBP and LBP. Notice that LBP tends to give tighter bounds (i.e.\ higher values) compared to IBP.}
    \label{fig:Full-example}
\end{figure*}

\section{BOUNDS FOR PROBABILISTIC SAFETY}\label{sec:methodology}
We show that the computation of $P_{\text{safe}}(T,S)$ reduces to computing the maximal set of weights for which the corresponding deterministic neural network is safe. 
To formalize this concept, consider the following definition.
\begin{definition}\label{Def:SafetyWeight}
 We say that ${H}\subseteq \mathbb{R}^{n_w}$ is the maximal safe set of weights from $T$ to $S$, or simply the maximal safe set of weights,  iff $H=\{ w \in  \mathbb{R}^{n_w}\, |\,\forall x \in T, f^{{w}}(x)\in S   \}.$ Furthermore, we say that $\hat{H}$ is a safe set of weights from $T$ to $S$, or simply a safe set of weights, iff $\hat{H} \subseteq H. $
\end{definition}
If $\hat{H}$ is a safe set of weights, then Definition \ref{Def:SafetyWeight} implies that for any  $w\in \hat{H}$  the corresponding neural network is safe, i.e., $\forall x \in T, f^w(x)\in S.$
Then, a trivial consequence of the definition of probabilistic safety is the following proposition.
\begin{proposition}
\label{Prop:SafetyComputation}
Let $H$ be the maximal safe set of weights from $T$ to $S$. Assume that 
$\mathbf{w}\sim q(w) $. Then, it holds that
\begin{align}
\label{Eq:IntegralSafety}
    \int_{H} q(w) dw = P_{\text{safe}}(T,S). 
\end{align} 
\end{proposition}
Proposition \ref{Prop:SafetyComputation} simply translates the safety property from the function space to an integral computation on the weight space.
As a consequence of Proposition \ref{Prop:SafetyComputation}, in the case of $q(w) = \mathcal{N}(w | \mu,\Sigma)$ with diagonal covariance, i.e., of posterior distribution computed by diagonal Gaussian VI, we obtain the following corollary.
\begin{corollary}
\label{Corollary:SafetyComp}
Assume that $\Sigma$, the covariance matrix of the posterior distribution of the weights,  is diagonal with diagonal elements $\Sigma_1,...,\Sigma_{n_w}$. Let $\hat{H}_1,...,\hat{H}_M$ be $M$  safe sets of weights such that, for $i\in \{1,..,M \}$, $\hat{H}_i=[l^i_1,u^i_1]\times ...\times [l_{n_w}^i,u_{n_w}^i]$ and $\hat{H}_i \cap \hat{H}_j=\emptyset$ for $i\neq j$. Then, it holds that
\begin{align*}
  P_{\text{safe}}&(T,S)\geq \\
  &\sum_{i=1}^M \prod_{j=1}^{n_w} \frac{1}{2} \left( \text{erf}\left(   \frac{ \mu_j - l_j^i  }{ \sqrt{2 \Sigma_j } }  \right)-\text{erf}\left(  \frac{ \mu_j - u_j^i  }{ \sqrt{2 \Sigma_j } }\right) \right) . 
\end{align*}
\end{corollary}

Proposition \ref{Prop:SafetyComputation} and Corollary \ref{Corollary:SafetyComp} guarantee that the computation of $P_{\text{safe}}$ is 
equivalent to the characterization of $H$, the maximal safe set of weights, and a lower bound for $P_{\text{safe}}$ can be computed by considering the union of $M$ safe sets of weights.
In what follows, in Section \ref{sec:IBPandLBP}, we derive a framework to efficiently check if a given set of weights is safe. Then, in Section \ref{sec::Algorithm} we present a method to generate safe sets of weights, which will be integrated in an algorithm for the computation of $P_{\text{safe}}(S,T)$ by making use of Proposition \ref{Prop:SafetyComputation}.

\subsection{SAFETY COMPUTATION}\label{sec:IBPandLBP}

In this subsection we derive schemes for checking whether a given hyper-rectangle,  $\hat{H}$, in the weight space is such that $\hat{H} \subseteq H$, that is, for $S=\{y\in \mathbb{R}^{n_c} \, |  \, C_S y + d_S \geq 0, C_S \in \mathbb{R}^{n_S\times n_c},d_S\in \mathbb{R}^{n_S}   \}$, we want to check whether $
    C_S f^{w}(x) + d_S \geq 0 ,  \quad \forall x \in T, \; \forall w \in \hat{H}.$
This is equivalent to check:
\begin{equation}\label{eq:det_spec}
    \min_{w \in \hat{H}, x \in T}  C_S f^{w}(x) + d_S \geq 0.
\end{equation}
In the following we show how IBP (Section \ref{sec:IBP}) and LBP (Section \ref{sec:LBP}) can be used to find a lower bound on the solution of the problem posed by Equation \eqref{eq:det_spec}.
The basic principles behind the two methods are depicted in Figure \ref{fig:Full-example} for an illustrative one-dimensional case (IBP shown in plots (a)--(c), LBP in plots (d)--(f)).
Given a bounding box in the input of each BNN layer (plot (a), as for a deterministic NN), and an interval in the weight space, $\hat{H}$ (plot (b), which is due to the fact that the BNN has probabilistic weights),  IBP propagates the two bounding boxes, as detailed in Section \ref{sec:IBP}, to obtain a bounding box on the network output (plot (c)).
The process is then iterated for each layer.
In LBP, instead, the linear function that bounds the input at each layer (plot (d)) is combined with the weight space interval $\hat{H}$ (plot (e)) to obtain a linear bound on the layer output (plot (f)) as detailed in Section \ref{sec:LBP}.
Intuitively, as LBP allows for a linear bound, it mimics more closely the behaviour of the network, thus giving better bounds, albeit at an increased computational cost.
This is further investigated in the experiments in Section \ref{sec:results}.

Before discussing IBP and LBP in detail, we first introduce common notation for the rest of the section. 
We consider fully-connected networks of the form:\footnote{CNNs can be considered by applying the approach of \cite{zhang2018efficient}.}
\begin{align}
    z^{(0)} &= x \label{eq:nn1} \\
    \zeta^{(k+1)}_i &=  \sum_{j=1}^{n_k} W^{(k)}_{ij} z^{(k)}_j + b^{(k)}_i  \quad i = 0,\ldots,n_{k+1}    \label{eq:nn2}\\
    z^{(k)}_i &= \sigma(\zeta^{(k)}_i )  \qquad  \qquad \qquad i = 0,\ldots,n_{k} \label{eq:nn3}
\end{align}
for $k=1,\ldots, K$, where $K$ is the number of hidden layers, $\sigma(\cdot)$ is a pointwise activation function, $W^{(k)} \in \mathbb{R}^{n_k \times n_{k-1}} $ and  $b^{(k)} \in \mathbb{R}^{n_k}$ are the matrix of weights and vector of biases that correspond to the $k$th layer of the network and $n_k$ is the number of neurons in the $k$th hidden layer.
We write $W^{(k)}_{i:}$ for the vector comprising the elements from the $i$th row of $W^{(k)}$, and similarly $W^{(k)}_{:j}$ for that comprising the elements from the $j$th column.
 $\zeta^{(K+1)}$ represents the final output of the network (or the logit in the case of classification networks), that is, $\zeta^{(K+1)} = f^{w}(x)$.
We write $W^{(k),L}$ and $W^{(k),U}$ for the lower and upper bound induced by $\hat{H}$ for $W^{(k)}$ and $b^{(k),L}$ and $b^{(k),U}$ for those of  $b^{(k)}$, for $k=0,\ldots,K$.
Notice that $z^{(0)}$, $\zeta^{(k+1)}_i$ and $z^{(k)}_i$ are all functions of the input point $x$ and of the combined vector of weights $w = [ W^{(0)}, b^{(0)} ,\ldots  , W^{(K)}, b^{(K)}  ]$.
We omit the explicit dependency for simplicity of notation.
Finally, we remark that, as both the weights and the input vary in a given set, 
Equation \eqref{eq:nn2} defines a quadratic form.

\subsubsection{Interval Bound Propagation}\label{sec:IBP}
IBP has already been employed for fast certification of deterministic neural networks \cite{gowal2018effectiveness}.
For a deterministic network, the idea is to propagate the input box around $x$, i.e., $T = [x^L,x^U]$,\footnote{In the general case, in which $T$ is not a box already, we first compute a bounding box $R = [x^L,x^U]$ such that $T \subset R$, and propagate $R$ with IBP, which yields a worst-case analysis.} through the first layer, so as to find values $z^{(1),L}$ and $z^{(1),U}$ such that  $z^{(1)} \in [ z^{(1),L}, z^{(1),U}]$, and then iteratively propagate the bound through each consecutive layer for $k=1,\ldots,K$. The final box constraint in the output layer can then be used to check for the property of interest \cite{gowal2018effectiveness}.
The only adjustment needed in our setting (that is,  for the bounding of Equation \eqref{eq:det_spec}) is that at each layer we also need to propagate the interval on the weight matrix $[W^{(k),L},W^{(k),U}]$ and that on the bias vector $[b^{(k),L},b^{(k),U}]$. 
This can be done by noticing that the minimum and maximum of each term of the bi-linear form of Equation \eqref{eq:nn2}, that is, of each monomial $W^{(k)}_{ij} z^{(k)}_j$, lies in one of the four corners of the interval $[W^{(k),L}_{ij},W^{(k),U}_{ij}] \times [z^{(k),L}_j,z^{(k),U}_j]$, and by adding the minimum and maximum values respectively attained by $b^{(k)}_i$.
As in the deterministic case, interval propagation through the activation function proceeds by observing that generally employed activation functions are monotonic, which permits 
the application of Equation \eqref{eq:nn3} to the bounding interval.
This is summarised in the following proposition.
\begin{proposition}
\label{Prop:IBP}
Let $f^{w}(x)$ be the network defined by the set of Equations \eqref{eq:nn1}--\eqref{eq:nn3}, let for $k = 0,\ldots,K$:
\begin{align*}
    t_{ij}^{(k),L} = \min \{ W_{ij}^{(k),L} z_j^{(k),L} , W_{ij}^{(k),U} z_j^{(k),L} , \\ W_{ij}^{(k),L} z_j^{(k),U} , W_{ij}^{(k),U} z_j^{(k),U}       \}  \\
    t_{ij}^{(k),U} = \max \{ W_{ij}^{(k),L} z_j^{(k),L} , W_{ij}^{(k),U} z_j^{(k),L} , \\ W_{ij}^{(k),L} z_j^{(k),U} , W_{ij}^{(k),U} z_j^{(k),U}       \} 
\end{align*}
where $i=1,\ldots,n_{k+1}$, $j=1,\ldots,n_k$, and $z^{(k),L} = \sigma(\zeta^{(k),L})$, $z^{(k),U} = \sigma(\zeta^{(k),U})$ and:
\begin{align*}
    \zeta^{(k+1),L} =  \sum_j t_{:j}^{(k),L} + b^{(k),L} \\
    \zeta^{(k+1),U} = \sum_j t_{:j}^{(k),U} + b^{(k),U}.
\end{align*}
Then we have that $\forall x \in T$ and $\forall w \in \hat{H}$:
\begin{equation*}
    f^w(x) =  \zeta^{(K + 1)} \in \left[  \zeta^{(K+1),L} , \zeta^{(K+1),U} \right].
\end{equation*}
\end{proposition}
The proposition above, whose proof is in the Supplementary Material (found in \cite{wicker2020probabilistic}), yields a bounding box for the output of the neural network in $T$ and $\hat{H}$. 
This can be used directly to find a lower bound for Equation \eqref{eq:det_spec}, and hence checking whether $\hat{H}$ is a safe set of weights.
As noticed in \cite{gowal2018effectiveness} and discussed in the Supplementary Material, for BNNs a slightly improved IBP bound can be obtained by eliding the last layer with the linear formula of the safety specification of set $S$.

\subsubsection{Linear Bound Propagation}\label{sec:LBP}
We now discuss how LBP can be used to lower-bound the solution of Equation \eqref{eq:det_spec}, as an alternative to IBP.
In LBP, instead of propagating bounding boxes, one finds lower and upper Linear Bounding Functions (LBFs) for each layer and then propagates them through the network.
As the bounding function has an extra degree of freedom w.r.t.\ the bounding boxes obtained through IBP, LBP usually yields tighter bounds, though at an increased computational cost.
Since in deterministic networks non-linearity comes only from the activation functions, LBFs in the deterministic case are computed by bounding the activation functions, and propagating the bounds through the affine function that defines each layer. 

Similarly, in our setting, given $T$ in the input space and $\hat{H}$ for the first layer in the weight space, we start with the observation that LBFs can be obtained and propagated through commonly employed activation functions for Equation \eqref{eq:nn3}, as discussed in \cite{zhang2018efficient}.


\begin{lemma}
\label{lemmma:act_bound}
Let $f^{w}(x)$ be defined by  Equations \eqref{eq:nn1}--\eqref{eq:nn3}. For each hidden layer $k=1,\ldots,K$, consider a bounding box in the pre-activation function, i.e.\ such that $\zeta^{(k)}_i \in [ \zeta^{(k),L}_i ,\zeta^{(k),U}_i ]$ for $i=1,\ldots,n_k$. Then there exist coefficients $\alpha^{(k),L}_i$, $\beta^{(k),L}_i$, $\alpha^{(k),U}_i$ and  $\beta^{(k),U}_i$  of lower and upper LBFs on the activation function such that for all $\zeta^{(k)}_i \in  [ \zeta^{(k),L}_i ,\zeta^{(k),U}_i ]$ it holds that:
\begin{equation*}
    \alpha^{(k),L}_i \zeta^{(k)}_i + \beta^{(k),L}_i \leq \sigma(\zeta^{(k)}_i) \leq \alpha^{(k),U}_i \zeta^{(k)}_i + \beta^{(k),U}_i. 
\end{equation*}
\end{lemma}
The lower and upper LBFs can thus be minimised and maximised to propagate the bounds of $\zeta^{(k)}$ in order to compute a bounding interval $[z^{(k),L},z^{(k),U}]$ for $z^{(k)} =  \sigma(\zeta^{(k)})$.
Then, LBFs for the monomials of the bi-linear form  of Equation \eqref{eq:nn2} can be derived using McCormick's inequalities \cite{mccormick1976computability}: 
\begin{align}
     W_{ij}^{(k)} z_j^{(k)} \geq W_{ij}^{(k),L} z^{(k)}_j  +  W_{ij}^{(k)} z^{(k),L}_j - W_{ij}^{(k),L} z^{(k),L}_j \label{eq:mc1} \\
     W_{ij}^{(k)} z_j^{(k)} \leq W_{ij}^{(k),U} z^{(k)}_j  +  W_{ij}^{(k)} z^{(k),L}_j - W_{ij}^{(k),U} z^{(k),L}_j \label{eq:mc2}
\end{align}
for every $i=1,\ldots,n_k$, $j=1,\ldots,n_{k-1}$ and $k=1,\ldots,K$.
The bounds of Equations \eqref{eq:mc1}--\eqref{eq:mc2} can thus be used in Equation \eqref{eq:nn2} to obtain LBFs on  the pre-activation function of the following layer, i.e.\ $\zeta^{(k+1)}$.
The final linear bound can be obtained by iterating the application of Lemma \ref{lemmma:act_bound} and Equations \eqref{eq:mc1}--\eqref{eq:mc2}  through every layer.
This is summarised in the following proposition, which is proved in the Supplementary Material (see \cite{wicker2020probabilistic}) along with an explicit construction of the LBFs.
\begin{proposition}
\label{proposition:lbp}
Let $f^{w}(x)$ be the network defined by the set of Equations \eqref{eq:nn1}--\eqref{eq:nn3}. Then for every $k=0,\ldots,K$ there exist lower and upper LBFs on the pre-activation function of the form:
\begin{align*}
     &\zeta^{(k+1)}_i \geq \mu_i^{(k+1),L} \cdot x + \sum_{l = 0}^{k-1} \langle \nu_i^{(l,k+1),L} , W^{(l)}  \rangle + \\ &\nu_i^{(k,k+1),L} \cdot W^{(k)}_{i:} + \lambda_i^{(k+1),L} \quad \textrm{for} \; i=1,\ldots,n_{k+1}\\
     &\zeta^{(k+1)}_i \leq \mu_i^{(k+1),U} \cdot x + \sum_{l = 0}^{k-1} \langle \nu_i^{(l,k+1),U} , W^{(l)}  \rangle + \\ &\nu_i^{(k-1,k+1),U} \cdot W^{(k)}_{i:} + \lambda_i^{(k+1),U}  \quad \textrm{for} \; i=1,\ldots,n_{k+1}
\end{align*}
where  $\langle \cdot , \cdot  \rangle $ is the Frobenius product between matrices, $\cdot$ is the dot product between vectors, and the explicit formulas for the LBF coefficients, i.e.,  $\mu_i^{(k+1),L}$, $\nu_i^{(l,k+1),L}$, $\lambda_i^{(k+1),L}$, $\mu_i^{(k+1),U}$, $\nu_i^{(l,k+1),U}$, are given in the Supplementary Material \cite{wicker2020probabilistic}.

Now let $\zeta^{(k),L}_i$ and $\zeta^{(k),U}_i$  respectively be the minimum and the maximum of the right-hand side of the two equations above; then we have that $\forall x \in T$ and $\forall w \in \hat{H}$:
\begin{equation*}
    f^w(x) =  \zeta^{(K + 1)} \in \left[  \zeta^{(K+1),L} , \zeta^{(K+1),U} \right].
\end{equation*}
\end{proposition}
Again, while the lower and upper bounds on $f^w(x)$ computed by Proposition \ref{proposition:lbp} can be directly used to check whether $\hat{H}$ is a safe set of weights, an improved bound can be obtained by directly considering the linear constraint form of $S$ when computing the LBFs.
This is further described in the Supplementary Material \cite{wicker2020probabilistic}.
\vspace{0.2em}
\section{ALGORITHM}
\label{sec::Algorithm}
\begin{algorithm} \small
\algsetup{linenosize=\small}
\caption{Probabilistic Safety for BNNs}\label{alg:simplereachability}
\textbf{Input:} $T$ -- compact input region, $S$ -- safe set, $C_S, d_S$ -- output constraints, $f^{\mathbf{w}}$ -- BNN, $\mathbf{w}\sim \mathcal{N}(\cdot | \mu,\Sigma)$ -- weight posterior, $\Sigma$ assumed to be diagonal, $N$ -- number of samples, $\gamma$ -- weight margin.  \\
\textbf{Output:} Safe lower bound on $P_{\text{safe}}(T,S)$. \\
\vspace*{-0.35cm}
\begin{algorithmic}[1]
\STATE $\hat{H} \gets \emptyset$ \COMMENT{$\hat{H}$ is the set of safe weights}
\FOR {$i\gets0$ to $N$} 
    \STATE $w' \sim \mathcal{N}(\cdot | \mu,\Sigma)$
    \STATE $[w^L, w^U] \gets [w'-\gamma \text{diag}(\Sigma), w'+\gamma \text{diag}(\Sigma)]$
    \STATE $y^L, y^U \gets \texttt{Method}(f, T, [w^L, w^U]$) \COMMENT{IBP/LBP}
    \IF{$\text{CheckProperty}$($C_S, d_S, y^L, y^U$)}
        \STATE $ H \gets \hat{H} \bigcup \{[w^L, w^U]\}$
    \ENDIF
\ENDFOR
\STATE $\hat{H} \gets \text{MergeOverlappingRectangles}(\hat{H})$ 
\STATE $p \gets 0.0$ 
\IF{$ \hat{H} \neq \emptyset$}
    \FOR{$[w^L, w^U] \in \hat{H}$} 
        \STATE $p \gets p + \prod_{i=1}^{n_w} \frac{1}{2} \left( \text{erf}\big(   \frac{ \mu_i - w^L_i  }{ \sqrt{2 \Sigma_i } }  \big)-\text{erf}\big(  \frac{ \mu_i - w^U_i }{ \sqrt{2 \Sigma_i } }\big) \right)$
    \ENDFOR
\ENDIF
\STATE return $p$
\end{algorithmic}
\end{algorithm}
%
%
In Algorithm \ref{alg:simplereachability} we illustrate our framework for lower-bounding of $P_{\text{safe}}(T,S)$ under the simplifying assumption that $\Sigma$ is diagonal.
The computational complexity of this algorithm for both IBP and LBP is discussed in the Supplementary Material \cite{wicker2020probabilistic}. 
In lines $1$--$10$ the algorithm computes $\hat{H}$, which is a safe set weights whose union is a subset of $H$, the maximal safe set of weights (see Definition \ref{Def:ProbSafety}).
In general, $H$ is not a connected set.
Hence, we build $\hat{H}$ as a union of hyper-rectangles (line $7$), each of which is
safe.
Each candidate safe hyper-rectangle is generated as follows: we sample a weight realisation, $w$, from the weights posterior (line 3) and expand each of its dimensions by a value computed by multiplying the variance of each weight with a proportional factor $\gamma$, which we refer to as the \textit{weight margin}  (line $4$). 
The trade-off is that greater values of $\gamma$ will yield larger regions, and hence potentially a larger safe region $\hat{H}$, though the chances that a wide interval $[w^L,w^U]$ will fail the certification test of line $6$ are higher.
This process is repeated $N$ times to generate multiple safe weight rectangles in disconnected regions of  the weight space. 
Empirically, we found this heuristic strategy to yield good candidate sets.\footnote{Note that the algorithm output is a lower bound of $P_{\text{safe}}(T,S)$ independently of the heuristic used for building $\hat{H}$.} In line $4$--$7$ we check if the given hyper-rectangle is safe by using either IBP or LBP, and by checking the safety specification (line $6$) as detailed in the Supplementary Material.
All safe rectangles are added to $\hat{H}$ in line $7$. 
In line $10$ we merge overlapping rectangles generated in the main algorithm loop, to guarantee that $\hat{H}$ is a set of non-overlapping hyper-rectangles of weights (as for Corollary \ref{Corollary:SafetyComp}).
This is done simply by iterating over the previously computed safe rectangles, and by merging them iteratively over each dimension.
Finally, in lines $12$--$15$, by employing  Corollary \ref{Corollary:SafetyComp}, we compute a lower bound for $P_{\text{safe}}(S,T)$.
The following theorem, which is a straightforward consequence of Proposition \ref{Prop:SafetyComputation}, \ref{Prop:IBP}, \ref{proposition:lbp} and Corollary \ref{Corollary:SafetyComp}, guarantees that Algorithm \ref{alg:simplereachability} returns a certified lower bound for $P_{\text{safe}}(S,T)$.
\begin{theorem}
Let $p$ be as computed by Algorithm \ref{alg:simplereachability}. Then, it holds that
$$ p \leq P_{\text{safe}}(S,T).$$
\end{theorem}
In the general case, when a non-Gaussian distribution or full covariance matrix is employed (e.g. with SWAG \cite{maddox2019simple} or HMC) the only modifications to make to the algorithm are in line $4$, where an estimation of the weight variance can be used, and in line $14$, which needs to be computed with Monte Carlo techniques. In this case, the weights sampled during integration can be utilised in line $3$ of Algorithm \ref{alg:simplereachability}. We expect that different posterior approximation methods with non-diagonal distributions will yield different probabilistic safety profiles, but stress that our method and its computational complexity is independent of the inference method used.

\begin{myexam}
\label{ex:RunningResults}
Lower bounds for the property discussed in Example \ref{ex:Running} are depicted in Figure \ref{fig:Full-example} (rightmost plots). 
We analyse the influence of parameters involved in Algorithm \ref{alg:simplereachability} (the number of samples $N$ and the weight margin $\gamma$), 
and the use of either IBP or LBP. 
As expected, we obtain higher values (that is, a tighter bound) as the number of samples increases, as this implies a better chance of finding a safe weight interval, and we observe similar behaviour for $\gamma$.
Notice also that the bounds provided by LBP (bottom row) are always slightly more tight than those provided by IBP (top row).
\end{myexam}

\section{EXPERIMENTAL RESULTS}\label{sec:results}
We explore the empirical performance of our framework. 
We begin with an extended analysis of the regression setting introduced in Example \ref{ex:Running}. We then turn our attention to an airborne collision avoidance scenario (VCAS)  \cite{katz2017reluplex,wang2018formal}. 
Finally, we explore the scalability of our approach on the MNIST dataset, and show that we compute non-trivial lower bounds on probabilistic safety even for networks with over 1M  parameters.
\paragraph{Regression Task}

\begin{figure}[h]
    \centering
    \includegraphics[width=0.40\textwidth]{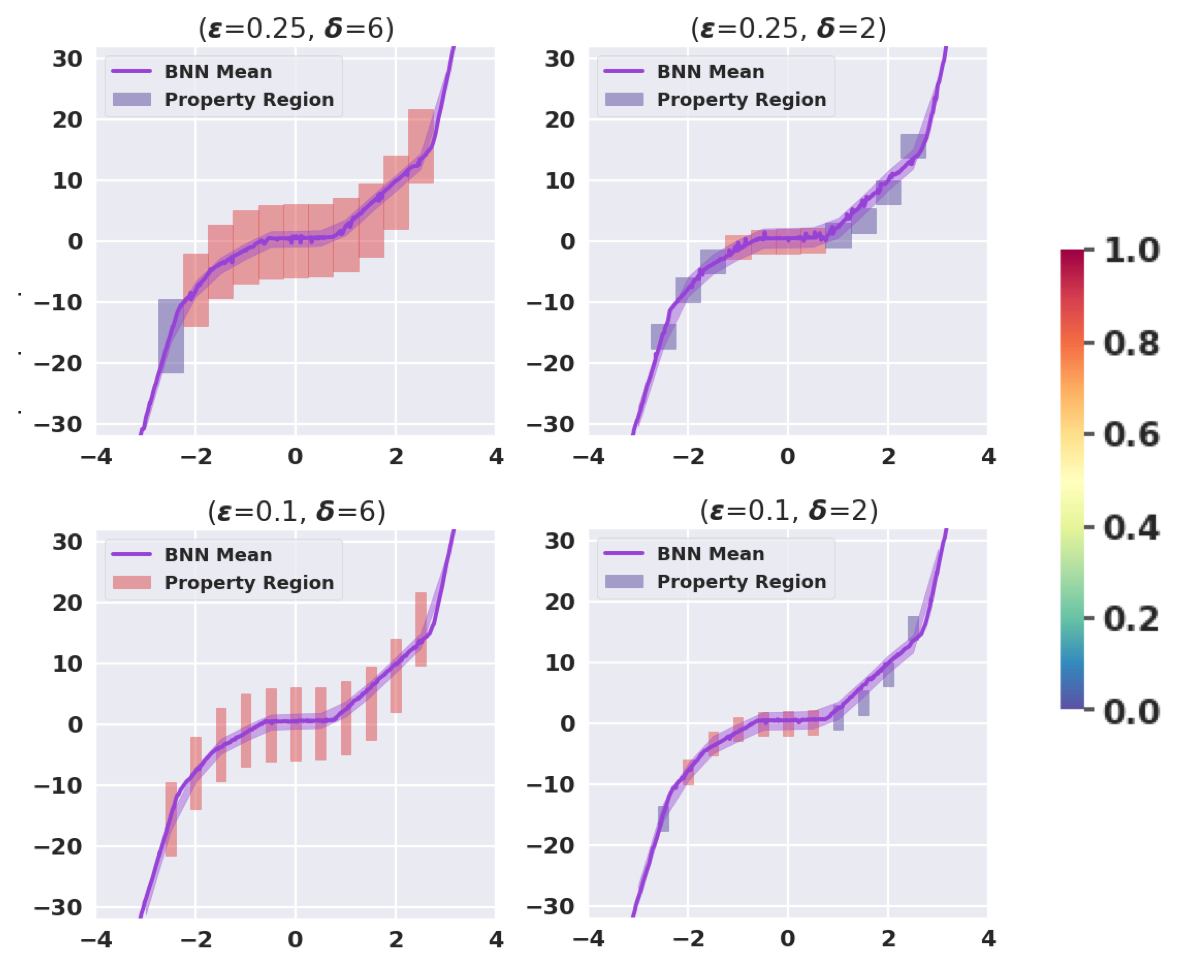}
    \caption{Probabilistic safety for the regression task on various input regions. Each box in the plot represents a safety specification and is colored with the lower bound returned by our method on the probability that the specification is met. The purple region represents 2 standard deviations about the mean of the BNN. }
    \label{fig:RegressionProp}
\end{figure}

In Figure~\ref{fig:RegressionProp} we explore the regression problem introduced in Example \ref{ex:Running}. 
We train a BNN with 128 hidden neurons for 10 epochs. 
We check the performance of our methodology on this BNN under different properties by considering many input regions $T$ and output sets $S$.
In particular, we extend the property in Example~\ref{ex:Running} to multiple intervals along the $x-$axis for different values of $\epsilon$ and $\delta$ and use LBP for the certification of safe weights regions.
Namely, we explore four different property specifications for the combination of $\epsilon \in \{0.1  , 0.25\}$ and $\delta  \in \{ 2  , 6 \}$, where every box in the plot represents both an input (range along the $x$ axis) and output region (range along the $y$ axis) and is colored accordingly with the lower bound obtained (which hence represents a lower bound on the probability that the samples from the BNN will remain inside that box for each specific range of $x$-axis values).
Naturally, we obtain higher values in regions in which the BNN is flat.
Also the bounds decreases as soon as  $\epsilon$ or $\delta$ increases as these imply a tighter property specification.

\paragraph{Airborne Collision Avoidance}
\begin{figure*}[h]
    \centering
    \includegraphics[width=0.75\textwidth]{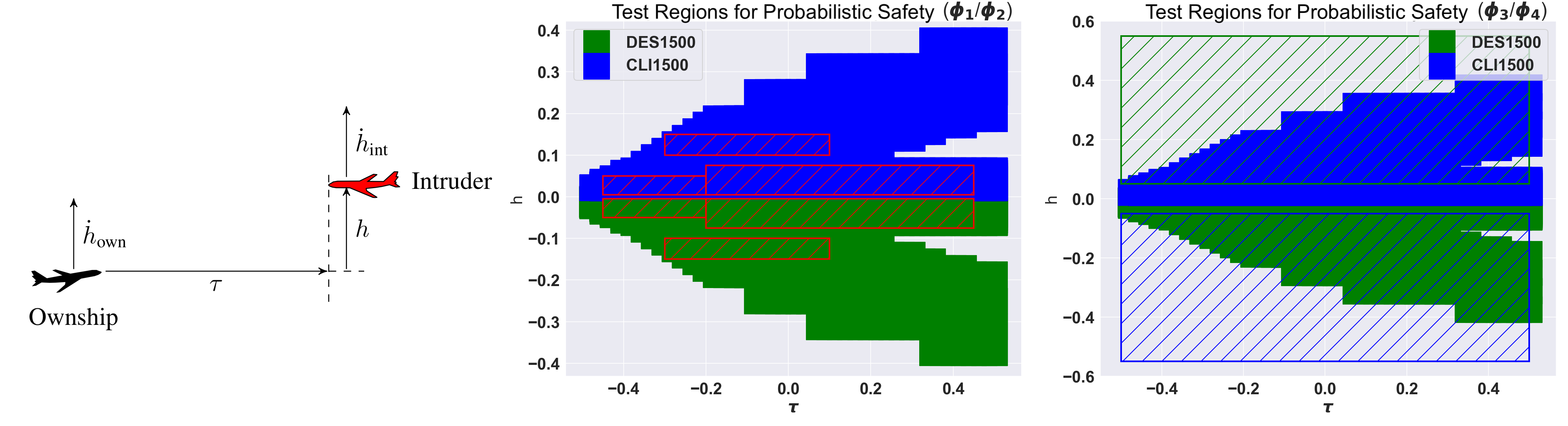}
    \caption{VCAS encounter geometry and properties under consideration. Left: Taken from \cite{julian2019guaranteeing}, a visualization of the encounter geometry and the four variables that describe it (distance $\tau$, ownship heading $\dot{h}_{\text{own}}$, intruder heading $\dot{h}_{\text{int}}$, and vertical separation $h$). Center: Visualization of ground truth labels (in color); red boxes indicate hyper-rectangles that make up the input areas for property $\phi_1$ (red boxes in the blue area) and $\phi_2$ (red boxes in the green area). Right: Hyper-rectangle for visualization of properties $\phi_3$ and $\phi_4$: we ensure that DES1500 is not predicted in the green striped box and CLI1500 is not predicted in the blue striped box.}
    \label{fig:VCAS-example}
\end{figure*}

We empirically evaluate probabilistic safety for the vertical collision avoidance system dataset (VCAS) \cite{julian2019guaranteeing}.
The task of the original NN  is to take as input the information about the geometric layout (heading, location, and speed) of the ownship and intruder, and return a warning if the ownship's current heading puts it on course for a near midair collision (NMAC). 
There are four input variables describing the scenario ( Figure~\ref{fig:VCAS-example}) and nine possible advisories corresponding to nine output dimensions. Each output is assigned a real-valued \textit{reward}. 
The maximum reward advisory indicates the safest warning given the current intruder geometry. 
The three most prominent advisories are clear of conflict (COC), descend at a rate $\geq$ 1500 ft/min (DES1500), and climb at a rate  $\geq$ 1500 ft/min (CLI1500). 
We train a BNN with one hidden layer with 512 hidden neurons  that  focuses on the situation in which the ownship's previous warning was COC, where we would like to predict if the intruder has moved into a position which requires action. This scenario is represented by roughly 5 million entries in the VCAS dataset and training our BNN with VI results in test accuracy of $91\%$.
We use probabilistic safety to evaluate whether the network is robust to four properties, referred to as $\phi_1$, $\phi_2$, $\phi_3$ and $\phi_4$,   
which comprise a probabilistic extension of those considered for NNs in \cite{katz2017reluplex, wang2018formal}.
Properties $\phi_1$ and $\phi_2$ test the consistency of DES1500 and CLI1500 advisories: given a region in the input space, $\phi_1$ and $\phi_2$ ensure that the output is constrained such that DES1500 and CLI1500 are the maximal advisories for all points in the region, respectively. 
On the other hand, $\phi_3$ and $\phi_4$ test that, given a hyper-rectangle in the input space, no point in the hyper-rectangle causes DES1500 or CLI1500 to be the maximal advisory.
The properties we test are depicted in the centre and right plot of Figure \ref{fig:VCAS-example}.
In Table ~\ref{tab:VCAS-Table} we report the results of the verification of the above properties, along with their computational times and the number of weights sampled for the verification.
Our implementation of Algorithm \ref{alg:simplereachability} with both  LBP and IBP is able to compute a lower bound for probabilistic safety with these properties in a few hundreds of seconds.\footnote{Note that in the case of $\phi_1$ and $\phi_2$ the input set $T$ is composed of three disjoint boxes. Our framework can be used on such sets by computing probabilistic safety on each box and then combining the results together via the union bound.}
Notice that, also in this case, LBP gives tighter bounds than IBP, though it takes around 5 times longer to run, which is a similar trade-off to what is observed for deterministic NNs \cite{zhang2018efficient}. 
%
\begin{table}\small
\centering
\resizebox{0.48\textwidth}{!}{%
 \begin{tabular}{||c | c c c c||} 
 \hline
 Method & Property & $P_{\text{safe}}$ & Time (s) & Num. Samples \\ [0.5ex] 
 \hline\hline
  \multirow{4}{*}{IBP}& $\phi_1$ & 0.9739 & 136 & 10500 \\ 
  &  $\phi_2$ & 0.9701 & 117 & 9000 \\
  &  $\phi_3$ & 0.9999 & 26 & 2000 \\
  &  $\phi_4$ & 0.9999 & 26 & 2000\\
\hline
\multirow{4}{*}{LBP}&$\phi_1$&0.9798 & 723 & 10500\\
     &$\phi_2$&  0.9867 & 628 & 9000\\
     &$\phi_3$& 0.9999 & 139 & 2000\\
     &$\phi_4$& 0.9999 & 148 & 2000\\
\hline
\end{tabular}}
\caption{\label{tab:VCAS-Table} VCAS probabilistic safety.
We see that, though LBP is ~5 times slower than IBP, it gives slightly tighter lower bounds, similarly to the observations in Figure~\ref{fig:Full-example}.
}
\end{table}
%
%
%
\paragraph{Image Classification on MNIST}
 We train several BNNs on MNIST to study how our method scales with the number of neurons, considering networks that are hundreds of times larger than those for the VCAS dataset.
 For this analysis we consider image classification with the MNIST dataset.
 In order to model  manipulations of an image we use the $l_\infty$-norm $\epsilon$-ball around test points. 
For all manipulations of magnitude up to $\epsilon$, 
we would like to check that the classification remains the same as that given by the (argmax of the) posterior predictive distribution. 
This can be done by first taking the classification according to the posterior on the test point $x^*$. Let $i$ be the predicted class index, 
then we create a $n_{\text{C}}\times n_{\text{C}}$ matrix, $C_{S}$, where the $i$th column is populated with ones and the diagonal is populated with negative ones, save for the $i$th entry which is set to one. Finally, ensuring that all entries of the vectors given by $C_{S}f^w(x)$ for all $x \in T$ are positive indicates that the classification does not change. 

\begin{figure}[h]
    \centering
    \includegraphics[width=0.45\textwidth]{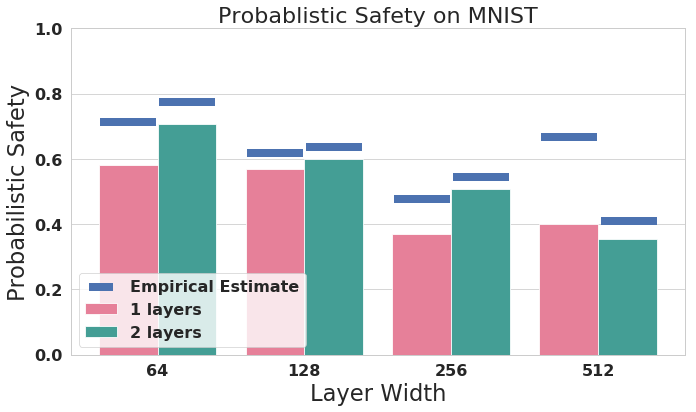}
    \caption{Performance of our framework on BNN architectures with varying numbers of neurons. The height of each bar represents the mean of the probabilistic safety value computed on 100 test set images. }
    \label{fig:MNIST-neurons}
\end{figure}

\paragraph{Evaluation.}
Using IBP, we are able to certify that the probabilistic safety of more than half of the tested 100 images is greater than 0.9 in the case of a 2 hidden layer BNN with 256 nodes per layer. In Figure~\ref{fig:MNIST-neurons}, we compare the lower bounds obtained with our approach with an empirical estimate obtained by sampling 500 posterior weights from the BNN posterior, so as to evaluate the tightness of the bounds obtained in practice. 
The results are given for the average over the same 100 images employed for the results reported in Figure ~\ref{fig:MNIST-neurons}. For 1 layer BNNs we use $\epsilon=0.025$ and for 2 layer BNNs we use $\epsilon=0.001$.
Notice that tight bounds can be obtained even for BNNs with almost a million of parameters, in particular, for a two-layer BNN with 512 neurons per layer, we have that our bound is within $5\%$ of the empirical results.



\section{CONCLUSION}
We considered probabilistic safety for BNNs, a worst-case probabilistic adversarial robustness measure, which can be used to certify a BNN against adversarial perturbations. 
We developed an algorithmic framework for the computation of probabilistic safety based on techniques from non-convex optimization, which computes a certified lower bound of the measure.
On experiments on various datasets we showed that our methods allows one to certify BNNs with general activation functions, multiple layers, and millions of parameters.

\paragraph{Acknowledgements}
This project was funded by the EU's Horizon 2020 research and innovation programme under the Marie Sk\l{}odowska-Curie (grant agreement No.~722022), the ERC under the European Union’s Horizon 2020 research and innovation programme (grant agreement No.~834115) and the EPSRC Programme Grant on Mobile Autonomy (EP/M019918/1).

\bibliographystyle{apalike}
\bibliography{bib}

\clearpage

\appendix 
\section{Relationship between Probabilistic Safety and other Measures}
In Proposition \ref{Prop:SafetyDefinition} we show that probabilistic safety, as defined in Definition \ref{Def:ProbSafety}, gives a lower bound to other measures commonly used to guarantee the absence of adversarial examples.
\begin{proposition}
\label{Prop:SafetyDefinition}
For $S\subseteq \mathbb{R}^{n_c}$ it holds that
$$ P_{safe}(T,S) \leq \inf_{x \in T} Prob(f^{\mathbf{w}}(x)\in S). $$
Moreover, if for $i \in \{1,...,n_c \}$, we assume that $S=\{y\in \mathbb{R}^{n_c}\, |\, y^{i}>a  \}$. Then, it holds that
$$ P_{safe}(T,S) \leq  \frac{\inf_{x \in T} \mathbb{E}_{w \sim {\mathbf{w}}}[ f^w_i(x)  ]}{a} .  $$
\end{proposition}
\paragraph{Proof of Proposition \ref{Prop:SafetyDefinition}}
\begin{align*}
    &  P_{safe}(T,S)=   \\
    &1 - Prob_{w \sim \mathbf{w}}(\exists x \in T, f^{w}(x) \in S)= \\
    &1- \mathbb{E}_{w \sim \mathbf{w}}[\sup_{x \in T} \mathbf{1}_{S}[f^w(x)  ] ]\leq \\
    &1 - \sup_{x \in T}\mathbb{E}_{w \sim \mathbf{w}}[\mathbf{1}_{S}[f^w(x)  ] ] =\\
&    1 - \sup_{x \in T}Prob_{w \sim \mathbf{w}}( f^w(x) \in S  )=\\
&\inf_{x \in T}Prob_{w \sim \mathbf{w}}( f^w(x) \in S  )=\\
&\inf_{x \in T}Prob_{w \sim \mathbf{w}}( f^w_i(x) \geq a  )\leq \\
&\frac{\inf_{x \in T} \mathbb{E}_{w \sim {\mathbf{w}}}[ f^w_i(x)  ]}{a},
\end{align*}
where the last inequality is due to Markov's inequality. 

\section{Computational Complexity}\label{sec:complexity}
Algorithm \ref{alg:simplereachability} has a complexity which is linear in the number of samples, $N$, taken from the posterior distribution of $w$.
The computational complexity of the method is then determined by the computational complexity of the method used to propagate a given interval $\hat{H}$ (that is, line 5 in Algorithm 1).
The cost of performing IBP is $\mathcal{O}(Knm)$ where $K$ is the number of hidden layers and $n \times m$ is the size of the largest weight matrix $W^{(k)}$, for $k=0,\ldots,K$. 
LBP is instead  $\mathcal{O}(K^2nm)$.

\section{Proofs}
In this section of the Supplementary Material we provide proofs for the main propositions stated in the paper.
\subsection{Proposition \ref{Prop:IBP}}
The bounding box can be computed iteratively in the number of hidden layers of the network, $K$.
We show how to compute the lower bound of the bounding box; the computation for the maximum is analogous. 

Consider the $k$-th network layer, for $k=0,\ldots,K$, we want to find for $i=1,\ldots n_{k+1}$:
\begin{equation*}
 \min_{ \substack{ W^{(k)}_{i:} \in [W^{(k),L}_{i:} , W^{(k),U}_{i:}]  \\ z^{(k)} \in [z^{(k),L},z^{(k),U}] \\ b^{(k)}_i \in [b^{(k),L}_i, b^{(k),U}_i ] }}    z^{(k+1)}_i = \sigma \left( \sum_{j=1}^{n_k} W^{(k)}_{ij} z^{(k)}_j + b^{(k)}_i  \right).
\end{equation*}
As the activation function $\sigma$ is monotonic, it suffice to find the minimum of:  $\sum_{j=1}^{n_k} W^{(k)}_{ij} z^{(k)}_j + b^{(k)}_i$.
Since $W^{(k)}_{ij} z^{(k)}_j$ is a bi-linear form defined on an hyper-rectangle, it follows that it obtains its minimum in one of the four corners of the rectangle $[W^{(k),L}_{ij} , W^{(k),U}_{ij}] \times [z^{(k),L}_j,z^{(k),U}_j]$.\\
Let  $t_{ij}^{(k),L} = \min \{ W_{ij}^{(k),L} z_j^{(k),L} , W_{ij}^{(k),U} z_j^{(k),L} , \\ W_{ij}^{(k),L} z_j^{(k),U} , W_{ij}^{(k),U} z_j^{(k),U} \}$
we hence have:
$$
\sum_{j=1}^{n_k} W^{(k)}_{ij} z^{(k)}_j + b^{(k)}_i \geq  \sum_{j=1}^{n_k} t_{ij}^{(k),L} + b^{(k),L}_i =: \zeta^{(k+1),L}_i.
$$
Thus for every $W^{(k)}_{i:} \in [W^{(k),L}_{i:} , W^{(k),U}_{i:}]$,  $z^{(k)} \in [z^{(k),L},z^{(k),U}]$ and $b^{(k)}_i \in [b^{(k),L}_i, b^{(k),U}_i ]$ we have:
$$
    \sigma \left( \sum_{j=1}^{n_k} W^{(k)}_{ij} z^{(k)}_j + b^{(k)}_i  \right) \geq \sigma \left( \zeta^{(k+1),L}_i \right) 
$$
that is $z^{(k+1),L}_i = \sigma \left( \zeta^{(k+1),L}_i \right)  $ is a lower bound to the solution of the minimisation problem posed above.


%
\subsection{Proposition \ref{proposition:lbp}}
We first state the following Lemma that follows directly from the definition of linear functions:
\begin{lemma}
\label{lemmma:linear_prop}
Let $f^L(t) = \sum_j a_j^L t_j + b^L $ and $f^U(t) = \sum_j a_j^U t_j + b^U$ be lower and upper  LBFs to a function $g(t)$ $\forall t \in \mathcal{T}$, i.e. $ f^L(t) \leq g(t) \leq  f^U(t) $ $\forall t \in \mathcal{T}$. Consider two real coefficients $\alpha \in \mathbb{R}$ and $\beta \in \mathbb{R}$.
Define
\begin{align}
    &\bar{a}_j^L = \begin{cases}  \alpha a_j^L \; \textrm{if} \,   \alpha \geq 0  \\ \alpha a_j^U \; \textrm{if} \,   \alpha < 0  \end{cases}
    \bar{b}^L = \begin{cases}  \alpha b^L + \beta \; \textrm{if} \,   \alpha \geq 0  \\ \alpha b^U  + \beta \; \textrm{if} \,  \alpha < 0  \end{cases} \label{eq:lin_transf1}\\
    &\bar{a}_j^U = \begin{cases}  \alpha a_j^U \; \textrm{if} \,   \alpha \geq 0  \\ \alpha a_j^L \; \textrm{if} \,   \alpha < 0  \end{cases}
    \bar{b}^U = \begin{cases}  \alpha b^U + \beta \; \textrm{if} \,   \alpha \geq 0  \\ \alpha b^L  + \beta \; \textrm{if} \,   \alpha < 0  \end{cases} \label{eq:lin_transf2}
\end{align}
Then:
\begin{align*}
\bar{f}^L(t) := \sum_j \bar{a}_j^L t_j + \bar{b}^L \leq \alpha g(t) + \beta \leq \sum_j \bar{a}_j^U t_j + \bar{b}^U \\ =: \bar{f}^U(t)
\end{align*}
That is, LBFs can be propagated through linear transformation by redefining the coefficients through Equations \eqref{eq:lin_transf1}--\eqref{eq:lin_transf2}.
\end{lemma}

We now proof Proposition  \ref{proposition:lbp} iteratively on $k=1,\ldots,K$ that is that for $i=1,\ldots,n_k$ there exist $f_i^{(k),L}(x,W)$ and $f_i^{(k),U}(x,W)$ lower and upper LBFs such that:
\begin{align}
        &\zeta^{(k)}_i \geq  f_i^{(k),L}(x,W) := \mu_i^{(k),L} \cdot x +  \label{eq:lbp1} \\ & \sum_{l = 0}^{k-2} \langle \nu_i^{(l,k),L} , W^{(l)}  \rangle + \nu_i^{(k-1,k),L} \cdot W^{(k-1)}_{i:} + \lambda_i^{(k),L}  \nonumber\\
     &\zeta^{(k)}_i \leq  f_i^{(k),U}(x,W) := \mu_i^{(k),U} \cdot x + \label{eq:lbp2} \\ & \sum_{l = 0}^{k-2} \langle \nu_i^{(l,k),U} , W^{(l)}  \rangle + \nu_i^{(k-1,k),U} \cdot W^{(k-1)}_{i:} + \lambda_i^{(k),U} \nonumber
\end{align}
and iteratively find valid values for the LBFs coefficients, i.e., $\mu_i^{(k),L}$, $\nu_i^{(l,k),L}$, $\lambda_i^{(k),L}$, $\mu_i^{(k),U}$, $\nu_i^{(l,k),U}$ and $\lambda_i^{(k),U}$.

For the first hidden-layer we have that $\zeta^{(1)}_i = \sum_j W_{ij}^{(0)} x_j + b^{(0)}_i$.
By inequality \eqref{eq:mc1} and using the lower bound for $b^{(0)}_i$ we have:
\begin{align*}
&\zeta^{(1)}_i \geq \sum_j \left( W_{ij}^{(0),L} x_j  +  W_{ij}^{(0)} x^{L}_j - W_{ij}^{(0),L} x^{L}_j  \right) + b^{(0),L}_i \\
&= W_{i:}^{(0),L} \cdot x + W_{i:}^{(0)} \cdot x^{L} - W_{i:}^{(0),L} \cdot x^{L} + b_i^{(0),L}
\end{align*}
which is a lower LBF on $\zeta^{(1)}$.
Similarly using Equation \eqref{eq:mc2} we obtain:
\begin{align*}
   \zeta^{(1)}_i \leq W_{i:}^{(0),U} \cdot x + W_{i:}^{(0)} \cdot x^{L} - W_{i:}^{(0),U} \cdot x^{L} + b_i^{(0),U}
\end{align*}
which is an upper LBF on $\zeta^{(1)}$. 
By setting: 
\begin{align*}
    \mu_i^{(1),L} &= W_{i:}^{(0),L} \quad , \quad  \mu_i^{(1),U} = W_{i:}^{(0),U} \\
    \nu_i^{(0,1),L} &= z^{(0),L} \quad , \quad  \nu_i^{(0,1),U} = x^{L}  \\
    \lambda_i^{(1),L} &= - W_{i:}^{(0),L} \cdot x^{L} + b_i^{(0),L}  \\  
    \lambda_i^{(1),U} &= - W_{i:}^{(0),U} \cdot x^{L} + b_i^{(0),U}
\end{align*}
we obtains LBFs $f_i^{(1),L}(x,W)$ and $f_i^{(1),U}(x,W)$ of the form \eqref{eq:lbp1}--\eqref{eq:lbp2}.

Given the validity of Equations \eqref{eq:lbp1}--\eqref{eq:lbp2} up to a certain $k$ we now show how to compute the LBF for layer $k+1$, that is given $f_i^{(k),L}(x,W)$ and $f_i^{(k),U}(x,W)$ we explicitly compute $f_i^{(k+1),L}(x,W)$ and $f_i^{(k+1),U}(x,W)$.
Let $\zeta^{(k),L}_i = \min f_i^{(k),L}(x,W) $ and $\zeta^{(k),U}_i = \max f_i^{(k),U}(x,W) $ the minimum and maximum of the two LBFs (that can be computed analytically as the functions are linear). 
For Lemma \ref{lemmma:act_bound} there exists a set of coefficients such that $ z^{(k)}_i = \sigma(\zeta^{(k)}_i) \geq \alpha^{(k),L}_i \zeta^{(k)}_i + \beta^{(k),L}_i$.
By Lemma \ref{lemmma:linear_prop} we know that there exists $\bar{f}_i^{(k),L}(x,W)$ with coefficients $\bar{\mu}_i^{(k),L}$, $\bar{\nu}_i^{(l,k),L}$, $\bar{\lambda}_i^{(k),L}$  obtained through Equations \ref{eq:lin_transf1}--\ref{eq:lin_transf2} such that:
\begin{align*}
    z^{(k)}_i  \geq \alpha^{(k),L}_i f_i^{(k),L}(x,W) + \beta^{(k),L}_i \geq \bar{f}_i^{(k),L}(x,W) 
\end{align*}
%
%
that is $\bar{f}_i^{(k),L}(x,W)$ is a lower LBF on $z^{(k)}_i$ with coefficients $\bar{\mu}_i^{(k),L}$, $\bar{\nu}_i^{(l,k),L}$, $\bar{\lambda}_i^{(k),L}$.
Analogously let $\bar{f}_i^{(k),U}(x,W)$ be the upper LBF on $z^{(k)}_i$ computed in a similar way.

Consider now the bi-linear layer $\zeta_i^{(k+1)} = \sum_{j} W_{ij}^{(k)} z_j^{(k)} + b_i^{(k)}$.
From Equation \eqref{eq:mc1} we know that: $W_{ij}^{(k)} z_j^{(k)} \geq  W_{ij}^{(k),L} z^{(k)}_j  +  W_{ij}^{(k)} z^{(k),L}_j - W_{ij}^{(k),L} z^{(k),L}_j$.
By applying Lemma \ref{lemmma:linear_prop} with $\alpha = W_{ij}^{(k),L}$ and $\beta = 0$ we know that there exists a lower LBF $\hat{f}_{ij}^{(k),L}(x,W)$ with  a set of  coefficients $a_{ij}^{(k),L}$, $b_{ij}^{(l,k),L}$ and  $c_{ij}^{(k),L}$ computed applying Equations \eqref{eq:lin_transf1}--\eqref{eq:lin_transf2} to $\bar{\mu}_i^{(k),L}$, $\bar{\nu}_i^{(l,k),L}$, $\bar{\lambda}_i^{(k),L}$ such that: $W_{ij}^{(k),L} z^{(k)}_j \geq \hat{f}_{ij}^{(k),L}(x,W)$.
Hence we have:
\begin{align*}
    & \zeta_i^{(k+1)} = \sum_j  W_{ij}^{(k)} z_j^{(k)} + b_i^{(k)} \geq \sum_j \big( W_{ij}^{(k),L} z^{(k)}_j  + \\
    & W_{ij}^{(k)} z^{(k),L}_j  - W_{ij}^{(k),L}  z^{(k),L}_j \big) + b_i^{(k),L} \geq  \\
    & \sum_j \hat{f}_{ij}^{(k),L}(x,W) +  \sum_j W_{ij}^{(k)} z^{(k),L}_j   - \\ & \sum_j W_{ij}^{(k),L}  z^{(k),L}_j + b_i^{(k),L} = \\
    & \sum_j  \big( a_{ij}^{(k),L} \cdot x +   \sum_{l = 0}^{k-2} \langle b_{ij}^{(l,k),L} , W^{(l)} \rangle \\ &  + b_{ij}^{kl-1,k),L} \cdot W^{(k-1)}_{j:} + c_{ij}^{(k),L} \big) + \\ & W_{i:}^{(k)} \cdot z^{(k),L} - W_{i:}^{(k),L}  z^{(k),L}.
\end{align*}
By setting 
\begin{align*}
&  \mu_i^{(k+1),L} = \sum_j a_{ij}^{(k),L}\\
&  \nu_i^{(l,k+1),L} = \sum_j b_{ij}^{(l,k),L}  \quad k = 0,\ldots,l-2 \\
&  \nu_i^{(k-1,k+1),L} = b_{i}^{(k-1,k),L} \\
&  \nu_i^{(k,k+1),L} =  z^{(k),L} \\
&  \lambda_i^{(k+1),L} = \sum_j c_{ij}^{(k),L} - W_{i:}^{(k),L} \cdot z^{(k),L} +  b_i^{(k),L} 
\end{align*}
and re-arranging the elements in the above inequality, we finally obtain:
\begin{align*}
    &\zeta_i^{(k+1)} \geq \mu_i^{(k+1),L} \cdot x + \sum_{l = 0}^{k-1} \langle \nu_i^{(l,k+1),L} , W^{(l)}  \rangle + \\
    &\nu_i^{(k,k+1),L} \cdot W^{(k)}_{i:} + \lambda_i^{k+1),L}  =: f_i^{(k + 1),L}(x,W)
\end{align*}
which is of the form of Equation \eqref{eq:lbp1} for the lower LBF for the $k+1$-th layer.
Similarly an upper LBF of the form of Equation \eqref{eq:lbp2} can be obtained by using Equation \eqref{eq:mc2} in the chain of inequalities above.

\section{Linear Specifications}
In this section of the Supplementary Material we discuss how the output of IBP and LBP can be used to check against specification of the form of Assumption \ref{assumption}, that is of the form:
\begin{align*}
    C_S f^w(x) + d_S \geq 0  \quad \forall x \in T \; \forall w \in \bar{H}
\end{align*}
with  $C_S \in \mathbb{R}^{n_S\times n_c}$ and $d_S\in \mathbb{R}^{n_S}$, $\bar{H} = [ w^L, w^U ]$ and $T = [ x^L, x^U ]$.
Let $i=1,\ldots,n_S$ then we need to check for every $i$ whether:  $C_{S,i} \cdot f^w(x) + d_{S,i} \geq 0$. 
Which is equivalent to compute
\begin{align}\label{eq:min_spec}
    \min_{x \in T w \in \bar{H}}    C_{S,i} \cdot f^w(x) + d_{S,i} = \sum_{j=1}^{n_c} C_{S,ij}  f^w_j(x) + d_{S,i} 
\end{align}
and checking whether that is greater or equal to zero or not.
As presented in the main paper, Propositions \ref{Prop:IBP} and \ref{proposition:lbp} return a bounding box for the final output. 
Though this bounding box can be directly used to compute the minimum in Equation \eqref{eq:min_spec} (as this entails simply the minimisation of a linear function on a rectangular space), tighter bounds can be obtained both for IBP and for LBP. This is described in the following two subsections.
\subsection{IBP}
For IBP we can do something similar to what is done in the case of IBP for deterministic NN \cite{gowal2018effectiveness}.
Instead of propagating the bounding box up until the very last layer to $f^w_i(x) = \zeta^{K+1}_i$, we stop at the last hidden activation function values $z^{(K)}_i$.
We thus have:
\begin{align*}
    &\sum_{j=1}^{n_c} C_{S,ij}  f^w_j(x) + d_{S,i}  = \sum_{j=1}^{n_c} C_{S,ij}  \zeta^{(K+1)}_j + d_{S,i}  = \\
    &\sum_{j=1}^{n_c} C_{S,ij} \sum_{l=1}^{n_K} \left( W_{jl}^{(K)} z^{(K)}_l + b^{(K)}_j \right) + d_{S,i}  = \\
    & \sum_{l=1}^{n_K} \left( \sum_{j=1}^{n_c} C_{S,ij} W_{jl}^{(K)} \right) z^{(K)}_l + \sum_{j=1}^{n_c} C_{S,ij} b^{(K)}_j +    d_{S,i}.
\end{align*}
Notice that $\sum_{j=1}^{n_c} C_{S,ij} W_{jl}^{(K)}$ and $\sum_{j=1}^{n_c} C_{S,ij} b^{(K)}_j $ are linear transformation, of the weights and biases.
The lower and upper bounds on $W^{(K)}$ and $b^{(K)}$ can hence be propagated through this two functions to obtain lower and upper bounds that account for the specification $W^{(K),L}_S$, $W^{(K),U}_S$, $b^{(K),L}_S$ and $b^{(K),U}_S$.
Propagating that interval through the layer:
\begin{align*}
    \min_{\substack{z^{(K)} \in [ z^{(K),L} , z^{(K),U} ] \\ W^{(K)}_S \in [ W^{(K),L}_S , W^{(K),U}_S ] \\ b^{(K)}_S \in [ b^{(K),L}_S , b^{(K),U}_S ]  }  } W^{(K)}_S \cdot z^{(K)} + b^{(K)}_S
\end{align*}
gives a solution to Equation \eqref{eq:min_spec}.
\subsection{LBP}
For LBP one can simply proceed by propagating the linear bound obtained.
In fact, Proposition \ref{proposition:lbp} yields an upper and lower LBFs, $f^{(K+1),L}(x,w)$ and $f^{(K+1),U}(x,w)$ on $f^w(x)$ for every $x \in T$ and $w \in \hat{H}$. 
By Lemma \ref{lemmma:linear_prop} those two LBFs can simply be propagated through the linear specification of Equation \eqref{eq:min_spec} hence obtaining lower and upper LBFs on the full specification, which can then be minimised to checked for the validity of the interval $\hat{H}$. 

\section{Computational Resources}

All our experiments were conducted on a server equipped with two 24 core Intel Xenon 6252 processors and 256GB of RAM. For VCAS experiments no parallelization was necessary, whereas MNIST was parallelized over 25 concurrent threads.

\end{document}